\documentclass[10pt,twocolumn,letterpaper]{article}

\usepackage{cvm}
\usepackage{times}
\usepackage{epsfig}
\usepackage{graphicx}
\usepackage{amsmath}
\usepackage{amssymb}
\usepackage{multirow}
\usepackage{makecell}
\usepackage{amsmath}
\usepackage{algorithm}
\usepackage{algorithmic}
\usepackage{booktabs}

\usepackage[pagebackref=true,breaklinks=true,letterpaper=true,colorlinks,bookmarks=false]{hyperref}

\newcommand{\keywords}[1]{{\bf \emph{Keywords: #1}}}

\cvmfinalcopy
\def\cvmPaperID{61} 

\begin{document}

\title{Reinforced Label Denoising for Weakly-Supervised Audio-Visual Video Parsing}

\author{Yongbiao Gao$^{a,b,c,e}$, Xiangcheng Sun$^{a,b}$, Guohua Lv$^{a,b}$, Deng Yu$^{d}$, Sijiu Niu$^{e}$ 
\and $^a$\small Key Laboratory of Computing Power Network and Information Security, Ministry of Education, 
\and \small Shandong Computer Science Center (National Supercomputer Center in Jinan), 
\\ \small Qilu University of Technology (Shandong Academy of Sciences), Jinan, China
\\ $^b$\small Shandong Provincial Key Laboratory of Computing Power Internet and Service Computing, 
\\ \small Shandong Fundamental Research Center for Computer Science, Jinan, China 
\\ $^c$\small Key Laboratory of New Generation Artificial Intelligence Technology and Its Interdisciplinary Application (Southeast University), 
\\ \small Ministry of Education, China 
\\ $^d$\small School of Artificial Intelligence, Shandong University
\\ $^e$ \small Shandong Key Laboratory of Ubiquitous Intelligent Computing, Jinan, China}

\maketitle

\begin{abstract}
    Audio-visual video parsing (AVVP) aims to recognize audio and visual event labels with precise temporal boundaries, which is quite challenging since audio or visual modality might include only one event label with only the overall video labels available. Existing label denoising models often treat the denoising process as a separate preprocessing step, leading to a disconnect between label denoising and AVVP tasks. To bridge this gap, we present a novel joint reinforcement learning-based label denoising approach (RLLD). This approach enables simultaneous training of both label denoising and video parsing models through a joint optimization strategy. We introduce a novel AVVP-validation and soft inter-reward feedback mechanism that directly guides the learning of label denoising policy. Extensive experiments on AVVP tasks demonstrate the superior performance of our proposed method compared to label denoising techniques. Furthermore, by incorporating our label denoising method into other AVVP models, we find that it can further enhance parsing results.
\end{abstract}


\section{Introduction}
The way humans perceive knowledge is through multiple modalities, including auditory, visual, tactile, and taste. Numerous studies have demonstrated that integrating these various senses can enhance natural knowledge perception \cite{ref1,ref2,ref3,ref4}. Specifically, visual and auditory are particularly prevalent in conveying vast amounts of information during interactions with the environment.

In recent years, the majority of research focus on visual scene understanding \cite{ref22,ref23} or video parsing \cite{ref24,ref25}, often overlooking other modalities. Prior works in multi-modal learning \cite{ref26,ref27} primarily examine the synergy between text and visual modalities to acquire joint representations. However, auditory and visual data often coexist and provide complementary cues in human perception. Therefore, more recently, some works \cite{ref16,ref17,ref28,ref29,ref45,ref43,ref46} investigate on localizing audio and visual event tasks, called Audio-Visual Video Parsing (AVVP), which aims to generate temporal audio, video, and audio-visual event labels. Most of existing works explore weakly-supervised learning for AVVP task since densely annotated event modality and category label with temporal boundaries is extremely expensive and time-consuming. For instance, \cite{ref16} formulates the AVVP task as a Multimodal Multiple Instance Learning (MMIL) problem and introduces an attentive MMIL pooling method to adaptively explore useful audio and visual content from different temporal extents and modalities. While existing audio-visual parsing works often assume temporal alignment of audio and visual events, this alignment hypothesis may not hold in practical scenarios. As pointed out by \cite{ref18,ref19}, "some event clues that do not appear in one of the two modalities." This is exemplified by the saying "I have not seen the person, but I have heard his voice." For example, in a basketball game video, even if the commentator cannot be seen, their voice and cheering can be heard, making sound signals crucial for comprehensive video understanding. Thus, the commentator's speech serves as a noisy label for the visual modality, and the basketball is considered audio-specific noise in some video clips. This issue is defined as \textbf{modality-specific noisy}.
\begin{figure*}
    \centering
    \includegraphics[width=1\textwidth]{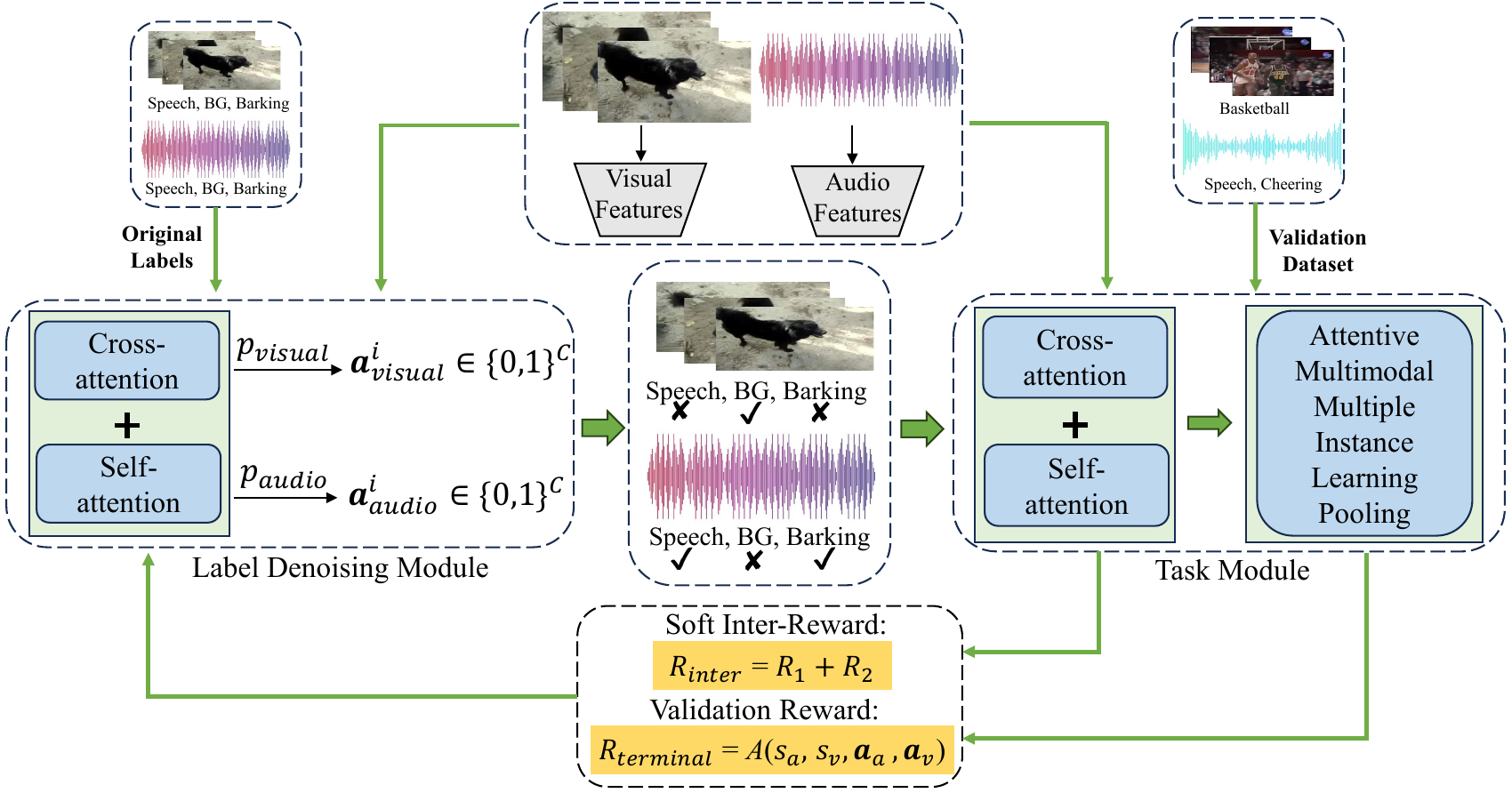}
    \caption{The overall framework of our proposed joint training RLLD for AVVP. The label denoising module aims to generate the denoising policy and the task module feedback the joint reward to guide the label denoising module learning.}
    \label{fig:enter-label}
\end{figure*}

Pioneering researchers in the field of audio-visual video parsing have made concerted efforts to address the modality-specific noisy problem. For example, an audio-visual modality exchange method \cite{ref18} aimed at deriving precise modality-aware event labels is proposed to, thus safeguard models from being misled by ambiguous overall labels. In a similar vein, \cite{ref19} tackle the modality-specific noisy issue by collaboratively computing a nosie estimator and forward modality losses. Leveraging the estimated noise ratios, they prioritize noisy samples based on forward losses. Notably, both the cross-modal and joint modal exchange mechanisms treat label denoising as a preliminary stage, disconnected from the subsequent AVVP process. This limitation precludes direct guidance of the denoising process by AVVP outcomes. Recently, VALOR \cite{ref40} incorporates large-scale contrastively pre-trained models as the modality teachers to enhance learning with weak labels. An iterative optimization approach \cite{ref43} using the EM algorithm is proposed to overcome the challenge of weak labels. However, the two most recent works ignore the modality-specific noisy problem. To address this gap, we introduce an innovative training strategy that integrates label denoising and audio-visual video parsing into a unified learning framework. This approach enables seamless interaction between denoising and parsing components, leading to more robust and accurate audio-visual video analysis.

In this paper, we introduce a novel Reinforcement Learning-based Label Denoising (RLLD) framework for audio-visual video parsing. The objective of RLLD is to enhance the identification of noisy labels and facilitate label denoising, thereby improving the overall performance of AVVP. Figure 1 provides an overview of the RLLD framework. The core component of RLLD is a reinforcement learning agent that is trained to identify noisy labels. This agent is guided by a joint reward signal derived from both the validation results and the soft inter-reward. This approach ensures that the denoising agent can maintain prediction accuracy while reducing convergence time. In addition, RLLD integrates label denoising and AVVP into a unified training framework. This integration enables the direct mitigation of label denoising and AVVP, facilitating more effective learning and optimization.

RLLD not only directly identifies noisy labels from a single modality, eliminating the need for computationally expensive cross-modal comparison, but also trains the denoising agent using validation results from the AVVP model. This approach ensures a more direct impact on enhancing AVVP performance, leading to superior temporal localization outcomes. We validate the effectiveness of RLLD on the standard AVVP dataset under several settings. The results demonstrate that RLLD achieves better performance than existing state-of-the-art label denoising models and further enhances the performance of state-of-the-art AVVP models. To summarize, our key contributions are as follows:\label{contributions}
\begin{itemize}
    \item We propose to address the label denoising issue in AVVP by reinforcement learning to learn the noisy label discovery strategy. To the best of our knowledge, we are the first to introduce reinforcement learning into AVVP to solve the label denoising problem.
    \item We carefully design the validation-based reward and the soft inter-reward functions to guide the label denoising strategy learning, which directly drives the label denoising for improving the performance of AVVP.
    \item The experiments on the AVVP dataset validate the effectiveness of our method, which can alleviate modality-specific noise during the jointly training framework.
\end{itemize}

\section{Related Work}
The traditional audio-visual learning primarily emphasize the representation learning \cite{ref8,ref7,ref9,ref6,ref5}, video captioning \cite{ref10,ref11,ref12}, audio-visual sound separation \cite{ref13,ref14,ref15}, {\it etc}. For example, \cite{ref9} propose a knowledge distillation-based procedure to transfer discriminative visual-knowledge from well established visual recognition models into the sound modality to get the better sound representation. Vid2Seq \cite{ref12} leverages unlabeled narrated videos for dense video captioning by reformulating sentence boundaries of transcribed speech as pseudo event boundaries, and using the transcribed speech sentences as pseudo event captions. All the past audio-visual related works assume audio and visual data are always correlated. In practice, the event label is contained in different modalities with different boundaries. Therefore, \cite{ref16} introduce a new fundamental problem: {\it audio-visual video parsing (AVVP)} that recognizes every event label containing in audio and visual modalities with temporal boundaries.

HAN \cite{ref16} is the pioneering weakly supervised framework to solve the audio-visual video parsing problem. HAN formulates the AVVP as a multi-instance learning problem and use a hybrid attention network to learn the unimodal and the cross-modal temporal contexts. The label smoothing technique is adopted to suppress the impact of noisy label from individual modality. Furthermore, MM-Pyramid \cite{ref17} attempt to capture temporal pyramid features via several stacking pyramid units. The previous proposed AVVP models take the overall event labels to supervise both visual and audio modalities learning. Recently, several latest models are proposed to solve the AVVP task like MGN \cite{ref45}, DHHN \cite{ref46}, CVCMS \cite{ref42}, CPSP \cite{ref41}, CMPAE \cite{ref50}, and poiBin \cite{ref43} etc. For example, audio, visual, and audio-visual streams are treated separately and an audio-visual class co-occurrence module is adopted to jointly explores the relationship of different categories among all streams \cite{ref42}. CPSP \cite{ref41} introduces the available full or weak label as a prior that constructs the exact positive-negative samples for contrastive learning. In practice, visual modality may not include the audio event, and vice versa. Therefore, the denoising methods \cite{ref18,ref19} aim to remove the noisy label for each modality by considering cross-modal connections and differences. LSLD \cite{ref48} designs language prompts to describe all cases of event appearance for each video since the language could freely describe how various events appear in each segment beyond fixed labels. Zhou \emph{et al.}\cite{ref49} use CLIP and CLAP to estimate the events in each video segment and generate segment-level visual and audio pseudo labels, respectively.
However, all the label denoising approaches in AVVP formulate the modality-specific denoising as a pre-processing stage, leading to a disconnect between label denoising and AVVP tasks.


In contrast to these methods, we generate reliable event labels independently for each modality via a joint training framework. We utilize validation results from the AVVP model to guide the learning of the label denoising strategy. This approach directly addresses the label denoising problem within the AVVP learning process, enabling more effective modality-specific denoising and improved overall performance.

\section{Method}
In this section, we elaborate on the new framework, Reinforcement Learning-based Label Denoising (RLLD) for AVVP. As depicted in Figure 1, RLLD comprises two distinct networks: label denoising network and task network. The label denoising network is implemented as a hybrid attention network for generating policy to determine which modality-specific label is the noisy label. We use the HAN \cite{ref16} as the task network to guide the denoising policy learning, though our method can not only adapt to HAN but also to any task networks for AVVP, such as \cite{ref40,ref43,ref45,ref46}, among others. 
\subsection{Problem Statement}
The input video sequence is denoted as $\{A_t,V_t\}^T_{t=1}$, where $T$ represents the overall duration of the video. Each $A_t$ and $V_t$ corresponds to the audio and visual modalities, respectively. The objective of AVVP is to segment the boundaries of audio and visual events and assign event labels that are present in each modality. During the training stage, only the overall video-label $\boldsymbol{y}\in\{0,1\}^C$ can be used to train the parsing network. Therefore, AVVP is a classic multiple instance learning problem. During validation stage, the audio, visual and audio-video event labels $\boldsymbol{y}_a^t\in\{0,1\}^C$, $\boldsymbol{y}_v^t\in\{0,1\}^C$, $\boldsymbol{y}_{av}^t\in\{0,1\}^C$ are available to assess whether the event occurs in both the visual and audio tracks at time $t$. In our formulation, following MA \cite{ref18} and JoMoLD \cite{ref19}, we aim to remove the modality-specific noisy label to improve the parsing performance. As a novel framework for label denoising in AVVP, RLLD differs from MA and JoMoLD in that we employ reinforcement learning to integrate label denoising with AVVP into a unified learning framework. This approach allows for joint optimization of denoising policy and AVVP task-specific objective.
\subsection{The Label Denoising Network}
We formulate the label denoising as a sequential decision-making process. Specifically, we develop a deep label denoising network that predicts probabilities for audio and visual tracks and determines which event label to remove from each modality based on the predicted probability distributions. We present a joint framework for training the label denoising network by using reinforcement learning.

Figure 1 visually illustrates how RLLD identifies noisy labels. A reinforcement learning agent is employed to map actions $a_t$ from the state $s_t$ to remove the event labels for each modality. We argue that the original event labels play a crucial role in identifying noisy labels. Therefore, the reinforcement learning state $s_t$ is defined as a combination of features and original noisy label representation, which is denoted as follows:
\begin{equation}
    \begin{aligned}
        s_{audio}^i = concat(f_{audio}^t,y_a^i),\\
        s_{visual}^i = concat(f_{visual}^t,y_v^i),\\
    \end{aligned}
\end{equation}
where $f_{a}^i$ and $f_{v}^i$ are audio and visual features extracted from pre-trained models. $i$ represents the index of the video.
After providing the audio and visual track states for a given video, following the approach of HAN \cite{ref16}, at each time step $t$, self-attention and cross-attention mechanisms are utilized to learn the audio and visual hidden state representations, respectively. This process is formalized as follows:
\begin{equation}
\begin{aligned}
    \hat{h}_{audio}^i &= s_{audio}^i + g_{sa}(s_{audio}^i,\boldsymbol{s}_{audio})+g_{ca}(s_{audio}^i,\boldsymbol{s}_{visual}), \\
    \hat{h}_{visual}^i &= s_{visual}^i + g_{sa}(s_{visual}^i,\boldsymbol{s}_{visual})+g_{ca}(s_{visual}^i,\boldsymbol{s}_{audio}),
\end{aligned}
\end{equation}
where $\boldsymbol{s}_{audio}=[s_{audio}^1;...;s_{audio}^T]$ and $\boldsymbol{s}_{visual}=[s_{visual}^1;...;s_{visual}^T]$. $g_{sa}(\cdot)$ and $g_{ca}(\cdot)$ are defined as follows:
\begin{equation}
    \begin{aligned}
        g_{sa}(s_{audio}^t,\boldsymbol{s}_{audio})&=softmax(\frac{s_{audio}^t\boldsymbol{s}_{audio}^{'}}{\sqrt{d}})\boldsymbol{s}_{audio},\\
        g_{ca}(s_{audio}^t,\boldsymbol{s}_{visual})&=softmax(\frac{s_{audio}^t\boldsymbol{s}_{visual}^{'}}{\sqrt{d}})\boldsymbol{s}_{visual},
    \end{aligned}
\end{equation}
where $d$ is the dimension of the audio and visual state. The fully connected ($FC$) layer, followed by the sigmoid function, is responsible for predicting the probability $p^t$ for each modality. Based on this probability, a modality-removal action $\boldsymbol{a}^t$ is sampled:
\begin{equation}
    \begin{aligned}
       p_{audio}^i &= \sigma(W\hat{h}_{audio}^i),\\
       p_{visual}^i &= \sigma(W\hat{h}_{visual}^i), \\
       \boldsymbol{a}_{audio}^i &\sim Bernoulli(p_{audio}^i),\\
       \boldsymbol{a}_{visual}^i &\sim Bernoulli(p_{visual}^i),
    \end{aligned}
\end{equation}
where $\sigma$ represents the sigmoid activate function, $\boldsymbol{a}_{audio}^i\in \{0,1\}^C$ and $\boldsymbol{a}_{visual}^i\in \{0,1\}^C$ sampled from Bernoulli distribution indicate whether the corresponding index event label is removing or not. Therefore, the revised label for each modality is obtained by:
\begin{equation}
    \begin{aligned}
       \hat{y}_a^i = \boldsymbol{a}_{audio}^i \odot y_a^i, \\
       \hat{y}_v^i = \boldsymbol{a}_{visual}^i \odot y_v^i, 
    \end{aligned}
\end{equation}
where $\odot$ indicates element-wise multiplication. $\hat{y}_a^i$ and $\hat{y}_v^i$ are the event labels for audio and visual tracks after removing noisy labels. 
\subsection{The Task Network}
During training, RLLD will receive a reward $R(S)$ that evaluates the quality of the denoising policy, and the objective of RLLD is to maximize the expected rewards over time by removing the noisy labels for each modality. In general, a high-quality label space is expected to obtain better performance in AVVP task. To this end, we propose a novel reward mechanism that directly feedback from the validation of the AVVP task.

We use the HAN as the basic task network with parameters $W$. We assume that the audio and visual snippets from temporal aggregated features are denoted as $\{\hat{f}_{audio}^t,\hat{f}_{visual}^t\}$. The aggregated features are predicted from self-attention and cross-attention networks similar to Eq.(2). Different from the label denoising network, the input of the task network only includes audio and visual features extracted from pre-trained models. The label representation is omitted in the task network. Subsequently, the attentive MMIL pooling is employed to predict video-level event probabilities. We denote the union of $\hat{y}_a^i$ and $\hat{y}_v^i$ as the ground-truth of the video-level label. 
\begin{equation}
    \hat{y} = \hat{y}_a^i\cup\hat{y}_v^i
\end{equation}
Therefore, the binary cross-entropy loss $CE(\cdot)$ is optimized to train the video-level prediction model.
\begin{equation}
    \mathcal{L}_{video} = -\sum_{c=1}^c \hat{y}[c]log(\Bar{p}[c])
\end{equation}
where $\Bar{p}$ is the predicted video-level event labels. Unlike the HAN approach, which employs label smoothing to mitigate modality bias, we utilize the outcomes from the label denoising network as the ground truth for both audio and visual modalities. The learning of audio and visual modalities is guided by the following loss function:
\begin{equation}
    \mathcal{L}_{a,v} = -(\sum_{c=1}^c \hat{y}_a[c]log(\Bar{p}_a[c])+\sum_{c=1}^c \hat{y}_v[c]log(\Bar{p}_v[c]))
\end{equation}
where the $\hat{y}_a$ and $\hat{y}_a$ are predicted audio and visual event labels from attentive MMIL pooling network.

Recognizing that validation accuracy serves as a direct indicator of the quality of the label denoising, we design a soft inter-reward to guide the learning of the label denoising policy at each step.
\begin{algorithm}
    \caption{RLLD algorithm}
    \textbf{Input}: The video sequential $\{A_t, V_t\}_{t=1}^T$, and the corresponding original category labels $y\in \{0,1\}^C$, max epoch $T_{max}$\\
    \textbf{Parameter}: The denoising networks parameters $\theta$, the task network parameters $W$\\
    \textbf{Output}: Network parameters $\theta$ and $W$ for denoising and task modules
    \begin{algorithmic}[1] 
        \STATE Let $t=0$.
        \WHILE{not convergent or $t \le T_{max}$}
        \STATE Generate action $\boldsymbol{a}_{audio}^i$ and $\boldsymbol{a}_{visual}^i$ by the denoising network;\\
        \STATE Removing the noisy labels according to Eq.(5) and get the inter-reward according to Eq.(9);\\
        \STATE Train the AVVP network using the Eq.(8) to obtain $W$.\\
        \ \ \ \ \ \ \ \ \ \ \ \ \ \ \ \ \ \ \ \ \ \ \            $W = W + \alpha\nabla_{W}\mathcal{L}_{a,v}$.
        \STATE Validate the AVVP model to obtain the reward $R(S)$ according to Eq.(11).\\
        \STATE Evaluate the gradients of the denoising network by Eq.(15),\\
        \ \ \ \ \ \ \ \ \ \ \ \ \ \ \ \ \ \ \ \ \ \ \            $\theta = \theta + \eta\nabla_{\theta}J(\theta)$. \\
        \STATE $t=t+1$
        \ENDWHILE
        \STATE \textbf{return} The denoising model and the AVVP model.
    \end{algorithmic}
    \end{algorithm}
\begin{eqnarray}
			\begin{aligned}
				& R_1=\exp \left(-\left(\sum_{j=1}^m l_{soft}^{(j)} \ln \frac{l_{soft}^{(j)}}{l_t^{i n_{(j)}}}+\sum_{j=1}^m l_t^{i n_{(j)}} \ln \frac{l_t^{i n_{(j)}}}{l_{soft}^{(j)}}\right) / 2\right) \\
				& R_2=\frac{\sum_{j=1}^m l_{soft}^{(j)} l_t^{i n_{(j)}}}{\sqrt{\sum_{j=1}^m\left(l_{soft}^{(j)}\right)^2} \sqrt{\sum_{j=1}^m\left(l_t^{i n_{(j)}}\right)^2}} \\
				& R_{inter}=\alpha_1 R_1+\alpha_2 R_2
			\end{aligned}
\end{eqnarray}
where $l_{soft}$ is the soft label from the original labels by a smoothing operation. $l_t$ indicates the revised labels at every time-step. The $R_{inter}$ gives an immediate reward after the each denoising action and provides a direction for the denoising policy learning. In addition, a validation-based reward is carefully designed to directly guides the label denoising for the improvement of AVVP.
\begin{equation}
    R_{terminal} = A(s_{a},s_{v},\boldsymbol{a}_{a},\boldsymbol{a}_{v})
\end{equation}
where $A(\cdot)$ represents the validation result on validation dataset $\mathcal{V}$. The final reward is combined by the soft inter-reward $R_{inter}$ and the terminal reward $R_{terminal}$.
\begin{equation}
    R(S) = R_{terminal} + R_{inter}
\end{equation}
Since $R$ is non-differentiable, we use policy gradient, described in the following section, to update the label denoising network. By utilizing the proposed framework, we successfully integrate label denoising and AVVP into a unified training paradigm. The outcomes of label denoising can be directly utilized to train the AVVP model, while the results of AVVP serve as direct guidance for learning label denoising strategies. This joint training approach enables complementary learning between the two components.

\subsection{Training with Policy Gradient}
The objective of label denoising for AVVP is to learn a policy function $\pi_\theta$ with parameters $\theta$ by maximizing the expected rewards:
\begin{equation}
    J(\theta)=\mathbb{E}_{p_\theta}(a_{1:N})[R(S)],
\end{equation}
where $a_{1:N}$ denotes the probability distributions over possible actions for each modality, and $R(S)$ is computed by Eq.(9). $\pi_\theta$ is defined by our proposed RLLD.

Due to the fact that rewards can only be obtained after training video parsing verification based on denoising model results, there is a problem with delayed rewards. Therefore, the REINFORCE algorithm \cite{ref20} is introduced to compute the derivate of the object $J(\theta)$, {\it w.r.t} the parameters $\theta$ as:
\begin{equation}
    \nabla_\theta J(\theta) = \mathbb{E}_{p_\theta}(a_{1:N})[R(S)\sum_{i=1}^N\nabla_\theta log\pi_\theta(a_i|h_i)],
\end{equation}
where $a_i$ corresponds to actions $\boldsymbol{a}_{audio}^i$ and $\boldsymbol{a}_{visual}^i$. $h_i$ corresponds to the hidden state representation $\hat{h}_{audio}^i$ and $\hat{h}_{visual}^i$. Therefore, the Eq.(11) can be denoted as follows:
\begin{equation}
\nabla_\theta J(\theta) = \mathbb{E}_{p_\theta}(a_{a}^{1:N},a_{v}^{1:N})[R(S)\sum_{i=1}^N\nabla_\theta log\pi_\theta(a_a^i,a_v^i|h_a^i,h_v^i)].
\end{equation}
Since Eq.(12) involves the expectation over high-dimensional action, which is hard to compute directly, a Monte Carlo approximation for the above quantity is:
\begin{equation}
\nabla_\theta J(\theta) \approx \frac{1}{N}\sum_{i=1}^N \nabla_\theta log\pi_\theta(a_a^i,a_v^i|h_a^i,h_v^i) R(S),
\end{equation}
where $N$ is the step length of one episode. Finally, we summarize our proposed RLLD approach for label denoising in AVVP in \textbf{Algorithm 1}.

\subsection{Discussion}
As a novel framework for AVVP, aimed at eliminating noisy labels in audio and visual modalities, RLLD stands out from previous methods. We initially integrate label denoising and AVVP into a unified framework. We employ reinforcement learning to explore effective label denoising strategies and leverage the feedback from AVVP to guide the learning of these strategies. Ultimately, there remains a crucial question regarding the motivation behind our proposed RLLD.

{\it Why do we utilize reinforcement learning to learn the label denoising policy for AVVP?} The primary reason is that the objective of label denoising in AVVP is to enhance video parsing performance. Reinforcement learning emphasizes goal-directed learning through interaction more than other machine learning paradigms \cite{ref21}. Therefore, the improvement in parsing accuracy can be directly aligned with the goal (reward) in reinforcement learning. Secondly, in practical scenarios, we often lack knowledge about which labels are noisy in audio and visual modalities. The inherent characteristic of reinforcement learning is exploratory learning. Therefore, we employ reinforcement learning and design a soft inter-reward to discover strategies for automatically identifying noise labels specific to each modality. Finally, the label denoising module generates the denoising policy, and the validation results from the AVVP module guide the learning of the denoising policy. This architecture seamlessly integrates the denoising module and AVVP module into a joint learning framework, significantly simplifying modality-specific label denoising for AVVP.
\begin{table*}[h]
    \centering
    { \begin{tabular}{c|ccccc|ccccc}
        \toprule
        \multirow{2}{*}{Methods}  &  \multicolumn{5}{c}{Segment-level} & \multicolumn{5}{|c}{Event-level} \\
       \cmidrule{2-6} \cmidrule{7-11}
        & A & V & AV & Type & Event & A & V & AV & Type & Event \\
         \midrule
         AVE  & 47.2 & 37.1 & 35.4 &39.9 &41.6 & 40.4 & 34.7 & 31.6 &35.5 & 36.5\\
         AVSDN & 47.8 & 52.0 & 37.1 & 45.7& 50.8 & 34.1 & 46.3 & 26.5& 35.6 & 37.7 \\
         HAN  & 60.1 & 52.9 & 48.9 & 54.0 & 55.4 & 51.3 & 48.9 & 43.0 & 47.7 & 48.0 \\
         MM-Paramid & 60.9 & 54.4 & 50.0 & 55.1 & 57.6& 52.7 & 51.8 & 44.4 & 49.9& 50.5 \\
         HAN+Co-teaching+  & 59.4 & 56.7 & 52.0 & 56.0 & 56.3& 50.7 & 53.9 & 46.6 & 50.4 & 48.7 \\
         HAN+JoCoR & 61.0 & 58.2 & 53.1 & 57.4 & 57.7 & 52.8 & 54.7 & 46.7 &51.4 & 50.3\\
         MGN & 60.8 & 55.4 & 50.4 & 55.5&57.2 & 51.1 & 52.4 & 44.4 & 49.3 & 49.1\\
         CVCMS & 59.2 & 59.9 & 53.4 & 57.5& 58.1 & 51.3 & 55.5 & 46.2 & 51.0 & 49.7\\
         DHHN  & 61.3 & 58.3 & 52.9 & 57.5& 58.1& 54.0 & 55.1 & 47.3 & 51.5 & 51.5\\
         \midrule
         MA  & 59.8 & 57.5 & 52.6 & 56.6 & 56.6& 52.1 & 57.4 & 45.8 & 50.8 & 49.4\\
         JoMoLD  & 60.6 & 62.2 & 56.0 &59.6 & 58.6& 53.1 & 58.9 & 49.4 & 53.8 & 51.4\\
         MA+CL  & 60.3 & 60.0 & 55.1 & 58.9& 57.9& 53.6 & 56.4 & 49.0 & 53.0& 50.6 \\
         JoMoLD+CL  & 61.3 & 63.8 & 57.2 &60.8 & 59.9& 53.9 & 59.9 & 49.6 & 54.5 & 52.5\\
         \midrule
          RLLD (ours) & 61.6 & 64.6 & 58.2 & 61.5 & 60.7 & 54.9 & 60.1 & 50.6 & 55.2 & 53.4\\
         RLLD+CL (ours) & \textbf{63.4} & \textbf{65.1} & \textbf{58.7} & \textbf{62.4}& \textbf{61.9} & \textbf{55.8} & \textbf{61.2} & \textbf{51.2} & \textbf{56.1} & \textbf{54.0}\\
         \midrule
         CPSP & 58.5 & 57.8 & 52.6 & 56.3& 55.8 & 51.6 & 54.0 & 46.5 & 50.7 & 49.9\\     
         CPSP + ours & \textbf{60.1} & \textbf{59.0} & \textbf{54.7} & \textbf{57.9} & \textbf{57.2} & \textbf{53.1} & \textbf{55.6} & \textbf{47.9} & \textbf{52.2} & \textbf{51.3}\\
         \midrule
         VALOR  & 61.8 & 65.9 & 58.4 & 62.0 & 61.5 & 55.4 & 62.6 & 52.2 & 56.7 & 54.2\\
         VALOR + ours & \textbf{62.2} & \textbf{66.7} & \textbf{59.3} & \textbf{62.7} & \textbf{62.4} & \textbf{55.7} & \textbf{63.1} & \textbf{53.7} & \textbf{57.5} & \textbf{54.9} \\
         \midrule        
    \end{tabular}}
     \caption{Audio-visual video parsing F-score results (\%) in comparison and the enhancing results with the state-of-the-art methods on the testing set of LLP.}
    \label{table1}
\end{table*}
\section{Experiments} \label{experiments}
To evaluate the performance of RLLD for AVVP, we conduct comprehensive experiments. In this section, we present the details of experiment. All experiments are conducted using the PyTorch framework on two NVIDIA GeForce RTX 4090 GPUs under Pytorch platform. 
\subsection{Experimental Settings} \label{ex_setting}
\textbf{Dataset and metrics.} The {\it Look, Listen, and Parsing} (LLP) dataset \cite{ref16} is utilized to evaluate our proposed method. This dataset is a weakly supervised audio-visual event dataset containing 11,849 videos across 25 categories. Diverse event labels are collected to annotate each video, including {\it human speaking, baby crying}, and {\it car running, etc}. Notably, over 7,000 videos contain multiple event labels. The LLP dataset partitions the entire dataset into training, validation, and testing subsets. The validation and testing subsets include both audio and visual events. The validation subset comprises 649 videos, while the testing dataset comprises 1,200 videos. In line with previous studies \cite{ref16,ref19,ref17}, we employ segment-level and event-level F-scores for all modalities as evaluation metrics. The mIoU threshold is set to 0.5 for computing event-level F-scores. The overall performance across all events is computed by considering the results of all audio and visual events, denoted as Event@Audio-Visual. Averaging the metrics for audio, visual, and audio-visual results provides the Type@Audio-Visual performance measure. 

\textbf{Implementation Details.} Drawing parallels with prior works in AVVP, we utilize pre-trained ResNet512 \cite{ref30} and 3D ResNet \cite{ref31} to extract visual features, while pre-trained VGGish \cite{ref32} is employed for audio feature extraction. The learning rate for the label denoising model is set to {\it 1e-4}. The hyper-parameters of $\alpha_1$ and $\alpha_2$ are 0.6 and 0.4. The hyper-parameters of the task model follow the JoMoLD configuration. To address the state transitions issue in the label denoising module, we randomly sample 128 samples as a batch and 1/4 of the samples are overlapped in the next state. For the audio-specific denoising branch, we utilize the mean F-score (*0.1) of segment-level under the audio type on the validation dataset as feedback to guide audio-specific label denoising. Similarly, the mean F-score (*0.1) of event-level under the visual type on the validation dataset is used to guide visual-specific label denoising for the visual branch. In instances where the sampled action removes all labels, we maintain the labels unchanged, providing a feedback reward of -1. We utilize Adam to optimize both the denoising and task models, ensuring their optimal performance.

\textbf{Comparison Methods.} In this study, we compare our proposed RLLD model with for state-of-the-art label denoising methods for AVVP: JoMoLD \cite{ref19}, MA \cite{ref18} and with the enhanced contrastive learning (CL) loss. These methods aim to address modality-specific noise labels. The HAN utilizes label smoothing to address modality-specific challenges. JoMoLD and MA collaboratively exchange audio and visual modalities to identify noise labels. Additionally, we select weakly-supervised video parsing AVE \cite{ref36}, MM-Paramid \cite{ref17}, HAN \cite{ref16} and AVSDN \cite{ref39} as baselines for comparison. To further demonstrate the efficacy of our approach, we compare it with several state-of-the-art audio-visual event parsing methods: HAN with Co-teaching+ \cite{ref37} and JoCoR \cite{ref38} strategies, MGN \cite{ref45}, DHHN \cite{ref46}, CVCMS \cite{ref42}, CPSP \cite{ref41}, and poiBin \cite{ref43}. The HAN utilizes label smoothing to address modality-specific challenges. JoMoLD and MA collaboratively exchange audio and visual modalities to identify noise labels. In our model, after removing modality-specific noise labels, the visual and audio-specific labels are directly utilized as ground truth for training the AVVP model. 
What needs to be pointed out is that our proposed module is a label denoising method. To validate the effectiveness of our method, we also conducted experiments by integrating our proposed label denoising method into two state-of-the-art (SOTA) methods, VALOR \cite{ref40} and CPSP \cite{ref41}. These two SOTA models were selected as the task network. We employed the denoising setting to train the models, demonstrating that our proposed label denoising method can further enhance the performance of the SOTA AVVP models.

\subsection{Results}
Table \ref{table1} presents the F-score results (\%) for audio-visual video parsing in comparison with the state-of-the-art methods. In each table, the best performance is highlighted in boldface. As can be observed from the prediction results, our proposed RLLD method outperforms the baseline methods on F-score evaluation metrics. Specifically, RLLD significantly outperforms AVSDN and AVE on audio event type, achieving 61.5\% under segment level and 53.4\% under event-level. From the results, it is noteworthy that our approach performs marginally better than the state-of-the-art DHHN \cite{ref46} on visual (64.6\% \emph{vs} 58.3\% and 60.1\% \emph{vs} 55.1\%) and on audio (61.6\% \emph{vs} 61.3\% and 54.9\% \emph{vs} 54.0\%). Our method outperforms HAN with two enhanced strategies (60.7\% \emph{vs} 57.7\% and 53.4\% \emph{vs} 50.3\%) on audio-visual event type. MA and JoMoLD are two label denoising methods designed to address the AVVP task. The performance of our method is also superior to the most comparable methods under the same setting (w/o contrastive learning (CL) loss). Based on the results, we can conclude that our method achieves overall state-of-the-art performance.

The superior performance of our proposed RLLD can be attributed to two primary reasons. Firstly, we formulate the problem of label denoising as an agent exploration decision-making process. This approach allows the agent to automatically learn the optimal policy for removing noise labels, rather than relying on direct comparison between audio and visual modalities. Secondly, we directly utilize the validation results as rewards to guide the learning of the label denoising policy. This design ensures that the proposed RLLD is directly optimized for enhancing parsing performance. The results clearly demonstrate that leveraging validation results is more effective in guiding the reinforcement learning agent to learn the optimal label denoising policy, ultimately leading to superior video parsing results.
\begin{figure*}[htbp] \label{f2}
	\centering
	\begin{minipage}{0.24\linewidth}
		\centering
		\includegraphics[width=0.9\linewidth]{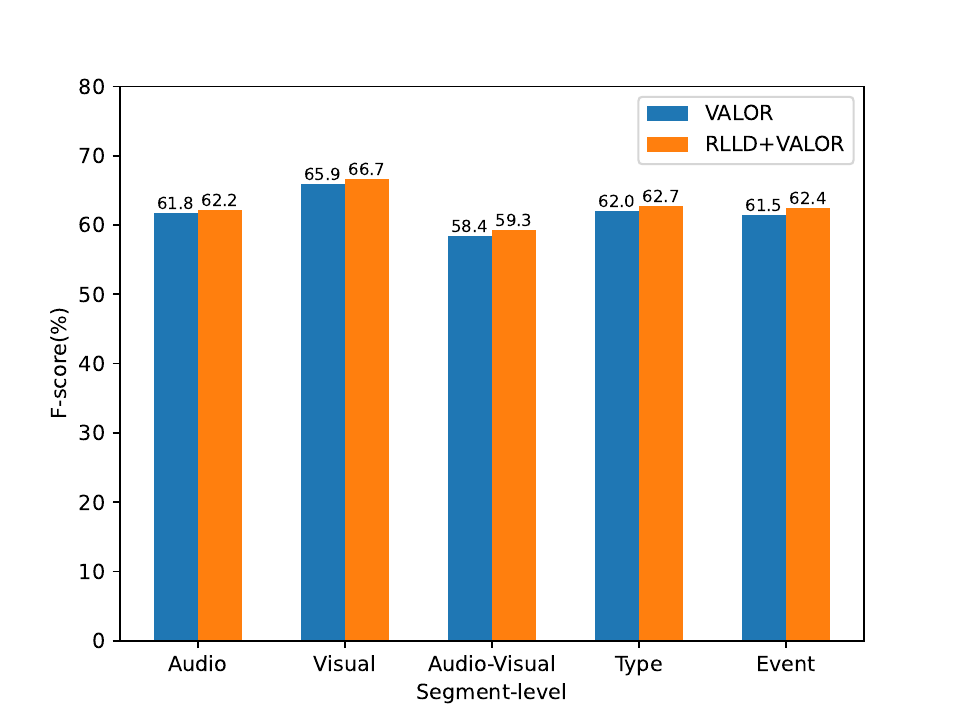}
		\caption{Segment-level results of VALOR and RLLD+VALOR.}
		\label{chutian1}
	\end{minipage}
	\begin{minipage}{0.24\linewidth}
		\centering
         \label{f3}
		\includegraphics[width=0.9\linewidth]{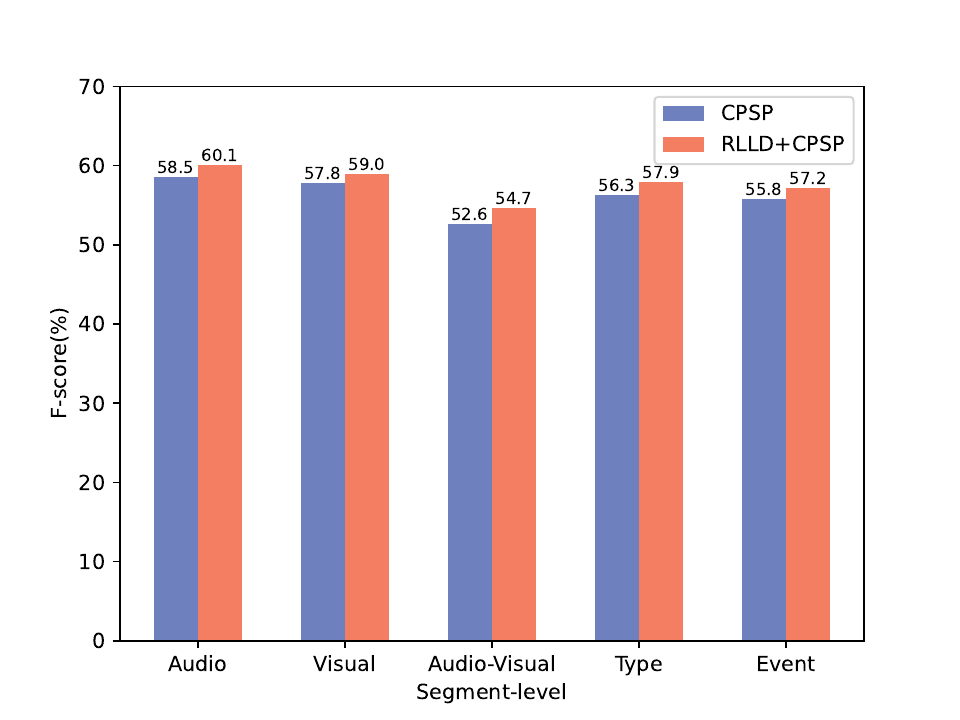}
		\caption{Segment-level results of CPSP and RLLD+CPSP.}
		\label{chutian2}
	\end{minipage}
        \begin{minipage}{0.24\linewidth}
		\centering
		\includegraphics[width=0.9\linewidth]{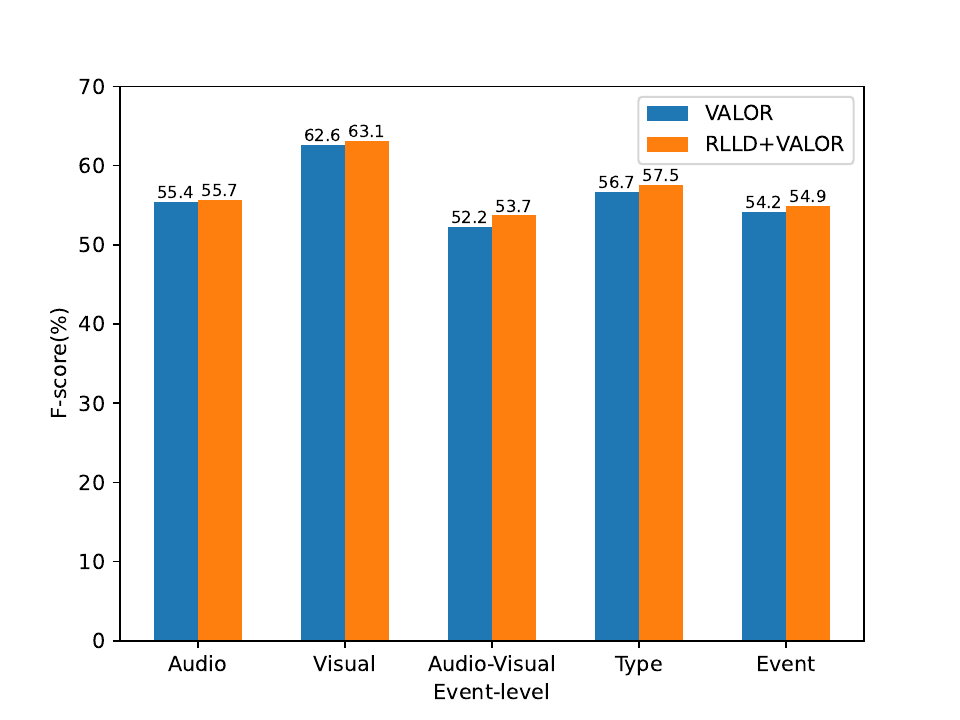}
		\caption{Event-level results of VALOR and RLLD+VALOR.}
		\label{chutian1}
	\end{minipage}
	\begin{minipage}{0.24\linewidth}
		\centering
		\includegraphics[width=0.9\linewidth]{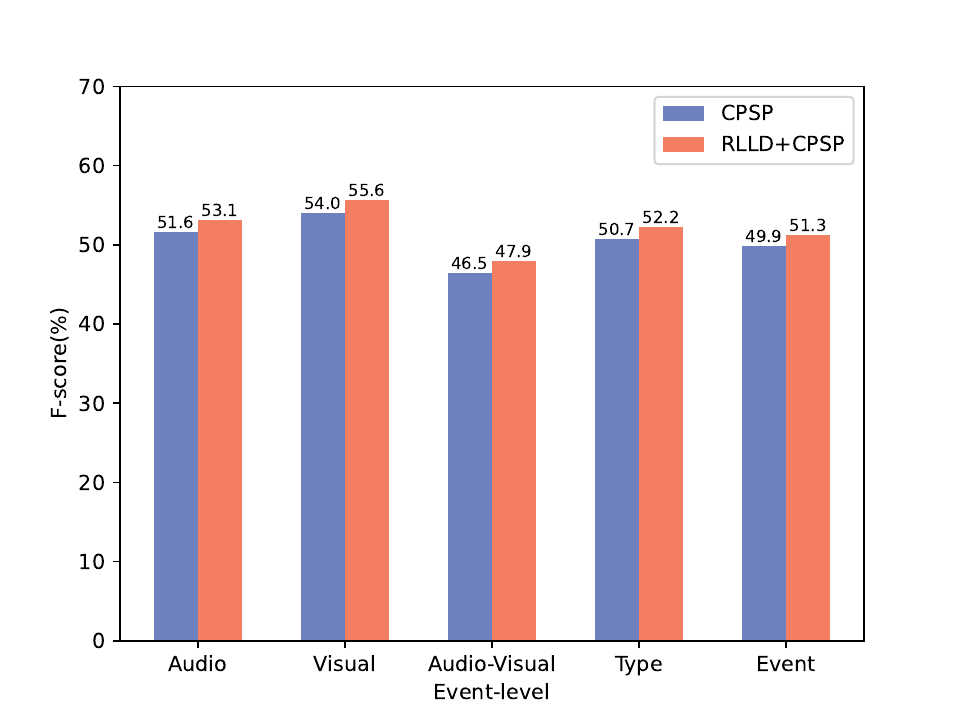}
		\caption{Event-level results of CPSP and RLLD+CPSP.}
		\label{chutian2}
	\end{minipage}
\end{figure*}
\begin{table*}
    \centering
   
    \centering
     \begin{tabular}{c|ccccc|ccccc}
        \toprule
        \multirow{2}{*}{Methods}  &  \multicolumn{5}{c}{Segment-level} & \multicolumn{5}{|c}{Event-level} \\
       \cmidrule{2-6} \cmidrule{7-11}
        & A & V & AV & Type & Event & A & V & AV & Type & Event \\
         \midrule
         RLLD w/o initialized labels  & 61.2 & 63.4 & 57.0 & 60.5&58.3 & 53.8 & 58.4 & 48.9 & 53.7 & 51.8\\
        RLLD w/o soft inter-reward   & 61.8 & 64.2 & 57.5 & 61.1& 60.1 & 54.6 & 59.3 & 50.1& 54.6 & 53.2 \\
         Full Setting (RLLD+CL) & \textbf{63.4} & \textbf{65.1} & \textbf{58.7} & \textbf{62.4} & \textbf{61.9} & \textbf{55.8} & \textbf{61.2} & \textbf{51.2} & \textbf{56.1} & \textbf{54.0} \\
        \midrule
    \end{tabular}
     \caption{Ablation results of the proposed modules on the LLP testing set.}
    \label{table1}
\end{table*}
\begin{figure}
	\centering
	\begin{minipage}{0.49\linewidth}
		\centering
		\includegraphics[width=0.9\linewidth]{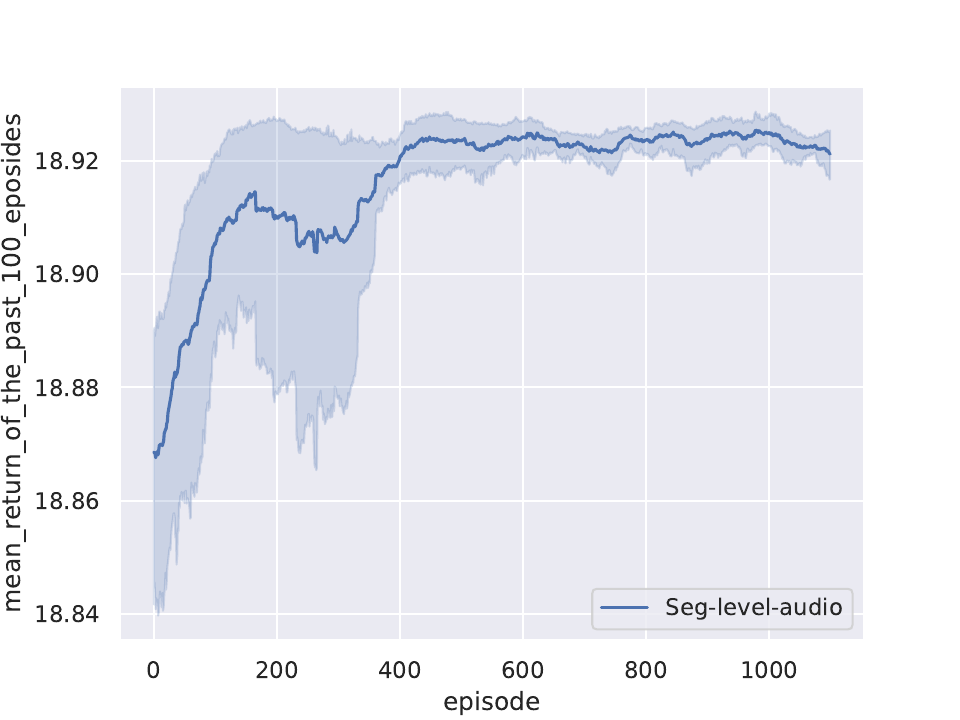}
		\caption{The audio curve under segment-level during the training.}
		\label{ccc}
	\end{minipage}
	\begin{minipage}{0.49\linewidth}
		\centering
		\includegraphics[width=0.9\linewidth]{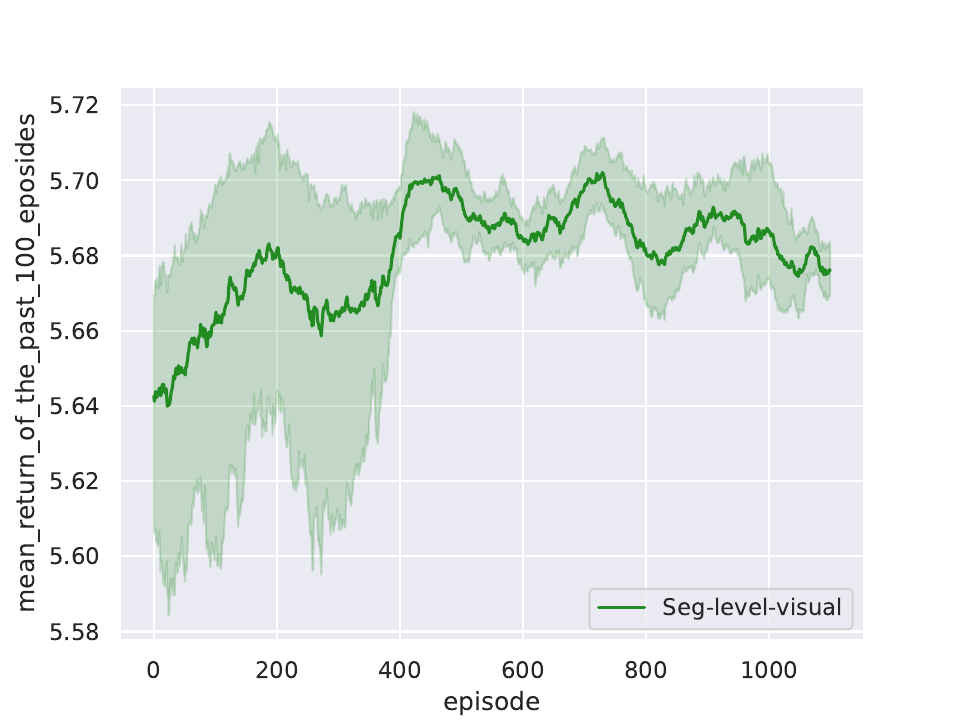}
		\caption{The visual curve under segment-level during the training.}
		\label{chutian2}
	\end{minipage}
\end{figure}

\begin{figure}
	\centering
	\begin{minipage}{0.49\linewidth}
		\centering
		\includegraphics[width=0.9\linewidth]{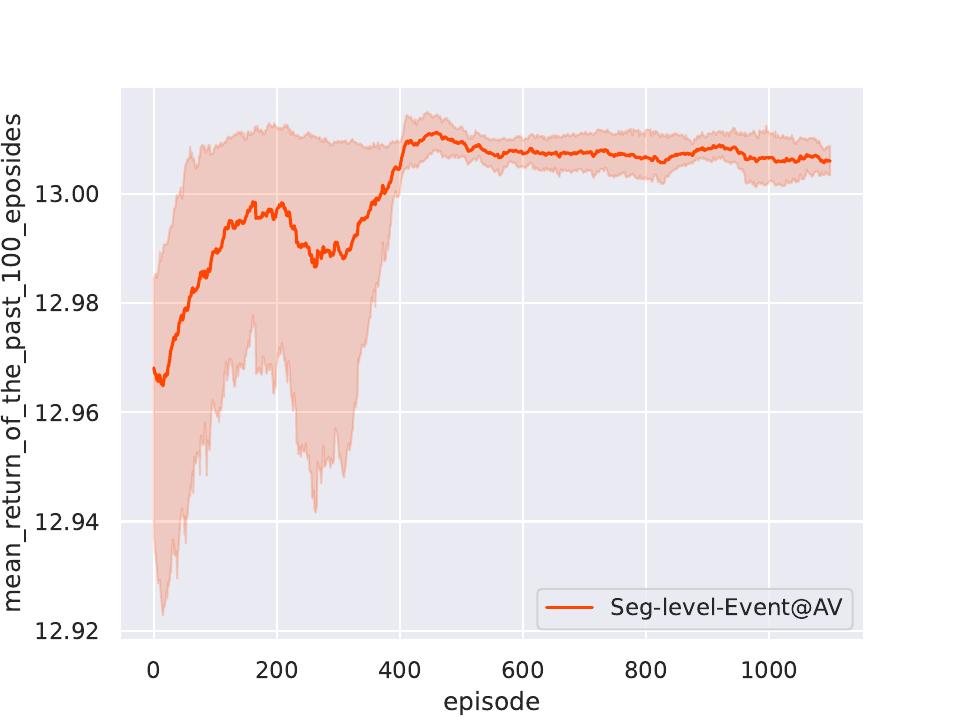}
		\caption{The event@AV curve under segment-level during the training.}
		\label{chutian1}
	\end{minipage}
	\begin{minipage}{0.49\linewidth}
		\centering
		\includegraphics[width=0.9\linewidth]{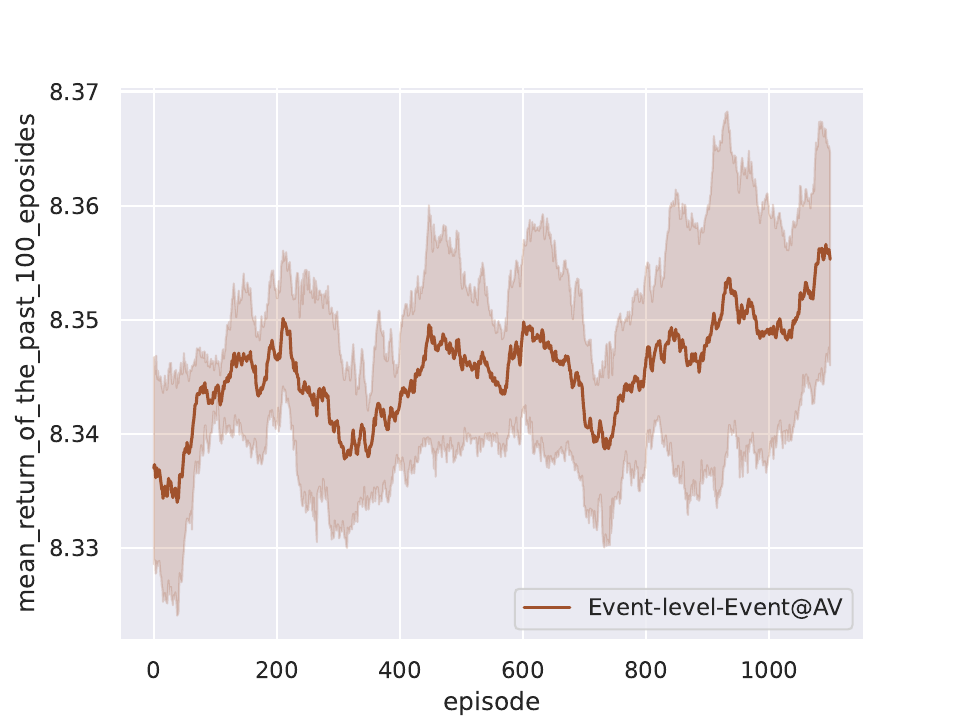}
		\caption{The event@AV curve under event-level during the training.}
		\label{chutian2}
	\end{minipage}
\end{figure}
\begin{figure}
	\centering
	\begin{minipage}{0.49\linewidth}
		\centering
		\includegraphics[width=0.9\linewidth]{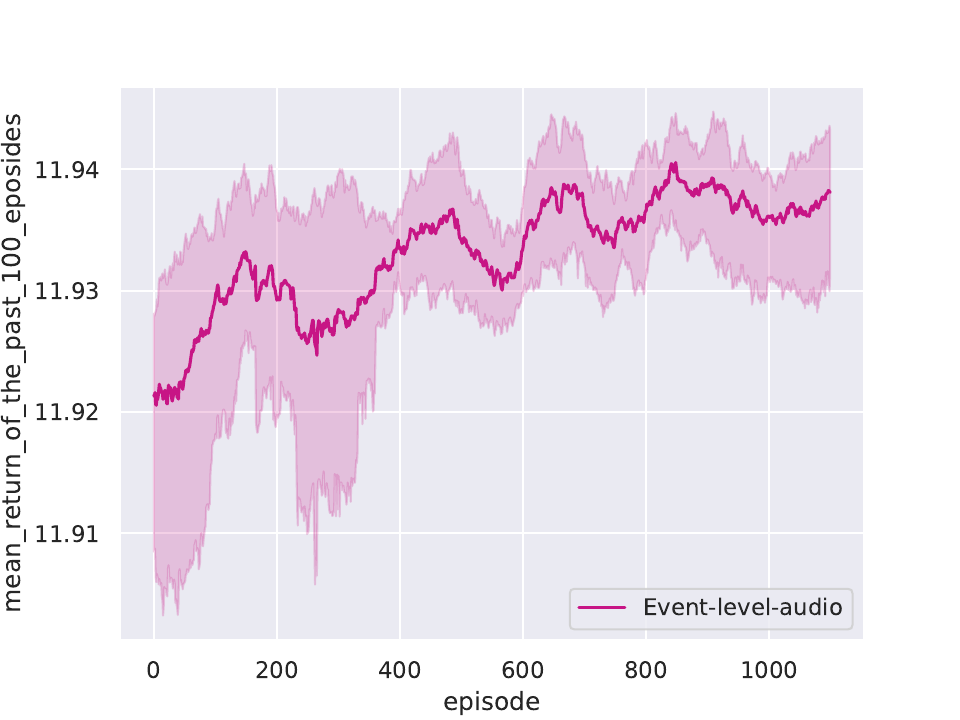}
		\caption{The audio curve under event-level during the training.}
		\label{chutian1}
	\end{minipage}
	\begin{minipage}{0.49\linewidth}
		\centering
		\includegraphics[width=0.9\linewidth]{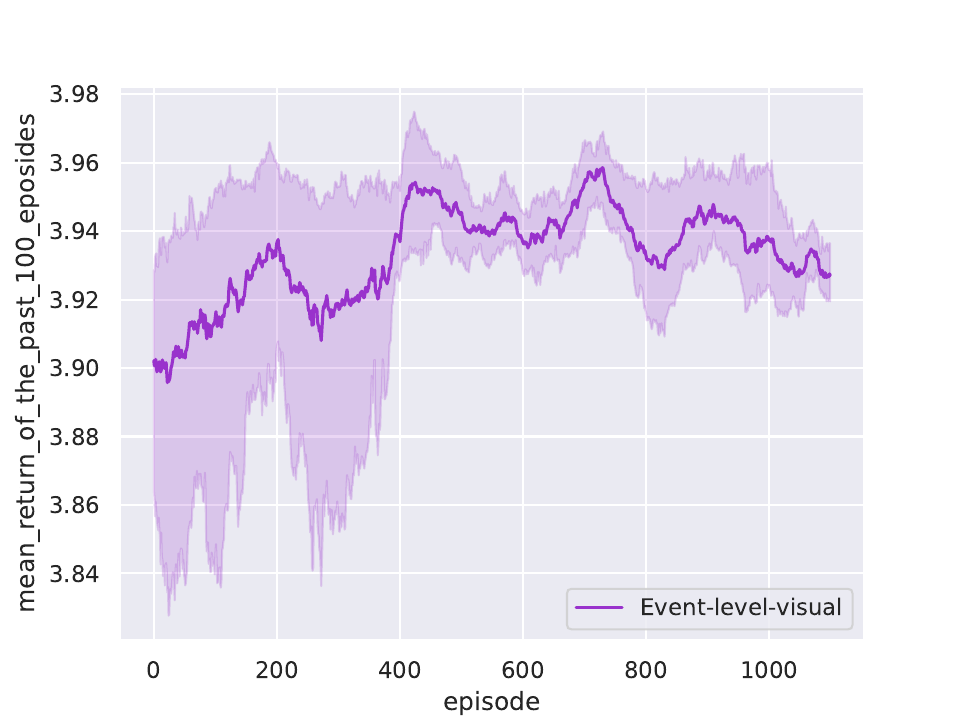}
		\caption{The visual curve under event-level during the training.}
		\label{chutian2}
	\end{minipage}
\end{figure}

\begin{table}[!h]
    \centering
    \centering
     \begin{tabular}{c|c|c}
        \toprule
        Methods  &  Parameters & training time (hours) \\
        \midrule
         MA  & 1.5*MMIL & $\approx 2.5$\\
        JoMoLD  & 2*MMIL & $\approx 3.0$\\
        \midrule
         Ours & 2*MMIL& $\approx 8.0$ \\
         \midrule
         \multicolumn{3}{l}{MMIL:The number of parameters in MMIL} \\ 
        \midrule
    \end{tabular}
     \caption{Comparison results of model parameters and training time.}
    \label{table1}
\end{table}
\subsection{Enhance the State-of-the-art Model}
Our proposed method is a label denoising technique that can further enhance any AVVP models. To demonstrate the capabilities of our label denoising approach, we integrated it into two state-of-the-art (SOTA) models: VALOR \cite{ref40} and CPSP \cite{ref41}.

For the 'RLLD+VALOR' setting, we maintain the pseudo-label extraction process from the pre-trained large model, eliminate the 'label elaboration' section in VALOR, and then, following Seq\_LE \cite{ref47}, select the top-k labels as relevant based on their probability values, ensuring that the sum of the top-k probabilities does not exceed 0.5. In this configuration, RLLD replaces the 'label elaboration' section in VALOR. For the 'RLLD+CPSP' setting, we retain the PSP and PSA models from CPSP and utilize the denoising results of RLLD to train CPSP during the classification stage. The quantitative analysis results are shown in the last two parts of Table 1, and the visualization results are shown in Figure 2-5. We can conclude that our proposed label denoising method effectively enhances the performance of the existing AVVP models. 

\subsection{Ablation Study}
The state representation of our proposed model incorporates both features and initialized labels. To assess the impact of each component within this representation, we conducted a rigorous ablation study by defining variations of RLLD. Specifically, we eliminated the initialized labels in the RLLD variant 'w/o initialized labels'. We conducted this ablation study on the LLP dataset, evaluating performance using the F-score metric. The results, summarized in Table 2, reveal that the ablated versions exhibit inferior performance compared to the complete RLLD. The ablation study highlights the crucial role of the components within the state representation.

Additionally, to evaluate the impact of the soft inter-reward, we conducted an ablation study that excluded it, focusing only on the terminal reward. The results of this experiment are also presented in Table 2. We find that removing the soft inter-reward led to a relatively significant decrease in experimental performance. Therefore, we can draw the conclusion that the soft inter-reward contributes significantly to enhancing the label denoising performance.

\subsection{Convergence Analysis}\label{convergence_description1}
To further demonstrate the convergence of our proposed RLLD, we present convergence curves on the LLP dataset under various metrics. Specifically, we calculate the F-score for the past 100 episodes between the predicted results and the ground truth of the validation dataset. To expedite the training process and reduce computational costs, we randomly select one-third of the training set as a subset for one-epoch model training. 

To obtain a representative convergence curve, we randomly initialize the model five times. The solid line represents the mean of these five experiments (0.1 smoothing), while the shaded area represents the corresponding standard deviation. The length of each episode is set to 32. The training process are shown in Figure 6-11. These figures reveal that the F-score for segment/Event audio metrics gradually increases with an increasing number of training episodes. For example, in the one-epoch validation feedback setting, the F-score of Segment-level@audio increases from 18.87 to 19.93. These convergence curves collectively demonstrate the convergence of our proposed method.

\subsection{Limitation Discussion} \label{limitation}
Reinforcement learning is a trial-and-error learning paradigm that necessitates a significant number of steps for the RL agent to explore the label denoising policy. The time consumption and model parameter quantity are shown in Table 3. Consequently, the proposed RLLD may incur a longer training duration to achieve optimal performance. It is worth noting that the high time complexity associated with reinforcement learning remains an open problem. Therefore, further investigations are necessary to address this issue and enhance the efficiency of RLLD.
\section{Conclusion}
In this paper, we introduce a novel label denoising method for AVVP. We formulate the label denoising task as a sequential decision-making process and introduce a validation task to steer the learning of the denoising policy. Our approach bridges the gap between label denoising and AVVP, providing a unified framework that directly addresses parsing enhancements in AVVP. We evaluate our proposed method on LLP datasets and showcase its effectiveness. An ablation study further verifies the significance of initialized labels and the meticulously designed soft inter-reward in the state space. Furthermore, we conduct experiments to demonstrate that our proposed label denoising method can further bolster AVVP models. Future work will concentrate on exploring enhanced reward mechanisms to improve policy learning for label denoising.
\section*{Acknowledgements}
This work was supported by the Natural Science Foundation of Shandong Province under Grant No. ZR2024QF115, the National Natural Science Foundation of China under Grant No. 62406155 and No. 62471202, Innovation Capability Enhancement Project for Technology-based Small and Medium-sized Enterprises of Shandong Province under Grant No. 2024TSGC0777, the Opening Fund of Shandong Provincial Key Laboratory of Ubiquitous Intelligent Computing, and the Development Program Project of Youth Innovation Team of Institutions of Higher Learning in Shandong Province.

{\small
\bibliographystyle{cvm}
\bibliography{cvmbib}
}

\end{document}